\documentclass{new_tlp}

\usepackage[utf8]{inputenc}
\usepackage{amsmath,amssymb}
\usepackage{tikz}
\usepackage{microtype}
\usepackage{listings}
\usepackage{bm}
\usepackage{multirow}
\usepackage{textcomp}
\usepackage{url}

\def\TPLP{}

\newtheorem{definition}{Definition}
\newtheorem{theorem}{Theorem}
\newtheorem{proposition}{Proposition}
\newtheorem{corollary}{Corollary}
\newtheorem{lemma}{Lemma}
\ifdefined\TPLP
\renewenvironment{proof}[1]{\paragraph{Proof~\ref{#1}.}}{$\Box$}
\def\qed{\hfill~$\boxtimes$}
\else

\fi

\newcommand{\Next}{\raisebox{-.5pt}{\Large\textopenbullet}} 
\newcommand{\previous}{\raisebox{-.5pt}{\Large\textbullet}} 
\newcommand{\wnext}{{\ensuremath{\widehat{\Next}}}}
\newcommand{\wprevious}{{\ensuremath{\widehat{\previous}}}}
\newcommand{\alwaysF}{\ensuremath{\square}}
\newcommand{\alwaysP}{\ensuremath{\blacksquare}}
\newcommand{\eventuallyF}{\ensuremath{\Diamond}}
\newcommand{\eventuallyP}{\ensuremath{\blacklozenge}}
\newcommand{\until}{\ensuremath{\mathbin{\bm{\mathsf{U}}}}}
\newcommand{\release}{\ensuremath{\mathbin{\bm{\mathsf{R}}}}}
\newcommand{\since}{\ensuremath{\mathbin{\bm{\mathsf{S}}}}}
\newcommand{\trigger}{\ensuremath{\mathbin{\bm{\mathsf{T}}}}}
\newcommand{\finally}{\ensuremath{\bm{\mathsf{F}}}}
\newcommand{\initially}{\ensuremath{\bm{\mathsf{I}}}}

\newcommand{\HT}{\ensuremath{\mathrm{HT}}}
\newcommand{\LTL}{\ensuremath{\mathrm{LTL}}}
\newcommand{\LTLf}{\ensuremath{\mathrm{LTL}_{\!f}}}
\newcommand{\LTLo}{\ensuremath{\mathrm{LTL}_{\omega}}}
\newcommand{\THT}{\ensuremath{\mathrm{THT}}}
\newcommand{\THTf}{\ensuremath{\mathrm{THT}_{\!f}}}
\newcommand{\THTo}{\ensuremath{\mathrm{THT}_{\!\omega}}}
\newcommand{\TEL}{\ensuremath{\mathrm{TEL}}}
\newcommand{\TELf}{\ensuremath{{\TEL}_{\!f}}}
\newcommand{\TELo}{\ensuremath{{\TEL}_{\omega}}}
\newcommand{\EM}{\ensuremath{\mathrm{EM}}}

\newcommand{\iequiv}{\ensuremath{\mathrel{\equiv_0}}} 
\newcommand{\gequiv}{\ensuremath{\mathrel{\equiv}}}   

\newcommand{\Head}[1]{\ensuremath{H(#1)}}

\newcommand{\Stamp}[2]{\ensuremath{#1_{#2}}} 
\newcommand{\initial}[1]{\ensuremath{\mathit{I}(#1)}} 
\newcommand{\dynamic}[1]{\ensuremath{\mathit{D}(#1)}} 
\newcommand{\final}[1]{\ensuremath{\mathit{F}(#1)}} 

\newcommand{\module}[1]{\ensuremath{\mathbb{#1}}}

\newcommand{\eqdef}{\ensuremath{\mathbin{\raisebox{-1pt}[-3pt][0pt]{$\stackrel{\mathit{def}}{=}$}}}}
\newcommand{\tuple}[1]{\langle #1 \rangle}
\renewcommand{\H}{\ensuremath{\mathbf{H}}}
\newcommand{\T}{\ensuremath{\mathbf{T}}}
\newcommand{\M}{\ensuremath{\tuple{\H,\T}}}

\newcommand{\loaded}{\ensuremath{\mathit{loaded}}}
\newcommand{\unloaded}{\ensuremath{\mathit{unloaded}}}

\newcommand{\shoot}{\ensuremath{\mathit{shoot}}}

\newcommand{\fail}{\ensuremath{\mathit{fail}}}
\newcommand{\goal}{\ensuremath{\mathit{goal}}}

\newcommand{\sysfont}{\textit}

\newcommand{\clingo}{\sysfont{clingo}}

\newcommand{\tel}{\sysfont{tel}}
\newcommand{\telingo}{\sysfont{telingo}}

\newcommand{\Bd}{\ensuremath{B}} 
\newcommand{\Hd}{\ensuremath{A}} 

\input{outline}

\title{Temporal Answer Set Programming on Finite Traces}

\author[Pedro Cabalar et al.]{%
  Pedro Cabalar\\
  University of Corunna, Spain
  \and
  Roland Kaminski, Torsten Schaub, Anna Schuhmann\\
  University of Potsdam, Germany}

\submitted{[n/a]}
\revised{[n/a]}
\accepted{[n/a]}

\begin{document}
\maketitle
%
\begin{abstract}
In this paper, we introduce an alternative approach to Temporal Answer Set Programming that relies on a variation of Temporal Equilibrium Logic (\TEL) for finite traces.
This approach allows us to even out the expressiveness of \TEL{} over infinite traces with the computational capacity of (incremental) Answer Set
Programming (ASP).
Also, we argue that finite traces are more natural when reasoning about action and change.
As a result, our approach is readily implementable via multi-shot ASP systems and benefits from an extension of ASP's full-fledged input language with
temporal operators.
This includes future as well as past operators whose combination offers a rich temporal modeling language.
For computation, we identify the class of temporal logic programs and prove that it constitutes a normal form for our approach.
Finally, we outline two implementations, a generic one and an extension of the ASP system \clingo.

\medskip\noindent
{\em Under consideration for publication in Theory and Practice of Logic Programming (TPLP)}
\end{abstract}
%

\section{Introduction}
\label{sec:introduction}

Representing and reasoning about dynamic systems is a key problem in Artificial Intelligence and beyond.
Accordingly, various formal systems have arisen,
including temporal logics~\cite{emerson90a} and calculi for reasoning about actions and change~\cite{sandewall94a}.
In Answer Set Programming (ASP;~\citeNP{lifschitz99b}),
this is reflected by
temporal extensions of Equilibrium Logic~\cite{agcadipevi13a},
the host logic of ASP,
and
action languages~\cite{gellif98a}.
Although both constitute the main directions of non-monotonic temporal systems,
their prevalence lags way behind the usage of plain ASP for modeling dynamic domains.
Hence,
notwithstanding the meticulous modeling of dynamics in ASP due to an explicit representation of time points,
it seems that its pragmatic advantages,
such as its rich (static) modeling language and readily available solvers,
often seem to outweigh the firm logical foundations of both dedicated approaches. 

Although the true reasons are arguably inscrutable,
let us discuss some possible causes.
The appeal of action languages lies in their elegant syntactic and semantic simplicity:
they usually consist of static and dynamic laws inducing a unique transition system.
Although most of them are implemented in ASP,
their simplicity denies the expressive possibilities of ASP.
Also, despite some recent reconciliation~\cite{leliya13a},
existing action languages lack the universality of ASP as reflected by the variety of variants.

Temporal Equilibrium Logic (\TEL;~\citeNP{agcadipevi13a}) builds upon
an extension of the logic of Here and There~(\HT;~\citeNP{heyting30a})
with Linear Temporal Logic (\LTL;~\citeNP{pnueli77a}).
This results in an expressive non-monotonic modal logic, which
extends traditional temporal logic programming approaches~\cite{cadivi15a} to the general syntax of \LTL{} and
possesses a computational complexity beyond \LTL~\cite{bozpea15a}.
As in \LTL, a model in \TEL{} is an \emph{infinite} sequence of states, called a trace.
This rules out computation by ASP technology (and necessitates model checking)
and
is unnatural for applications like planning,
where plans amount to finite prefixes of one or more traces.

Unlike this,
we address the representation and reasoning about dynamic systems by proposing
an alternative combination of the logics of \HT{} and \LTL{}
whose semantics rests upon finite traces.
On the one hand,
this amounts to a restriction of \TEL{} to finite traces.
On the other hand,
this is similar to the restriction of \LTL{} to \LTLf{} advocated by~\citeN{giavar13a}.
Our new approach, dubbed \TELf, has the following advantages.
First,
it is readily implementable via ASP.
Second,
it can be reduced to a normal form which is
close to logic programs and
much less complex than the one obtained for \TEL.
%
Finally,
its temporal models are finite and offer a one-to-one correspondence to plans.
Interestingly, \TELf{} also sheds light on concepts and methodology used in incremental ASP solving
when understanding incremental parameters as time points.

Another distinctive feature of \TELf{} is the inclusion of future as well as past temporal operators.
We associate this with the following benefits.
When using the causal reading of program rules,
it is generally more natural to draw upon the past in rule bodies and
to refer to the future in rule heads.
A similar argument was put forward by~\citeN{gabbay87a}.
This format also
yields a simpler normal form
and
lends itself to a systematic modeling methodology which favors the definition of states in terms of the past
rather than mixing in future operators.
For instance, in reasoning about actions,
the idea is to derive action effects for the current state and check their preconditions in the previous one,
rather than to represent this as a transition from the current to the next state.
This methodology aligns state constraints, effect axioms, etc.\ to capture the present state.
As well, past operators are much easier handled computationally than their future counterparts
when it comes to incremental reasoning, since they refer to already computed knowledge.

We make the above arguments more precise once our formal apparatus is set up.
In fact,
we introduce our approach in a more general semantic setting
encompassing not only \TELf{} but also its close ancestors \TEL, \LTL, and \LTLf.
This uniform base provides us with immediate insights into their interrelationships.
Once we have formally elaborated our approach (in the restricted space),
we turn to computational aspects.
First, we define the class of temporal logic programs and show that they constitute a normal form for \TELf.
Then, we provide bounded and time point-wise translations of temporal logic programs into regular ones.
Finally, we sketch two existing implementations:
\tel, computing bounded temporal models of \TELf{} theories, and
\telingo, incrementally computing temporal models of temporal logic programs over the extended full-fledged input language of \clingo.


\section{Temporal Equilibrium Logic on Finite Traces}
\label{sec:ftel}

All logics treated in this paper share the common syntax of \LTL{} with past operators~\cite{emerson90a}.
We start from a given set $\mathcal{A}$ of atoms which we call the \emph{alphabet}.
Then, a (temporal) \emph{formula} $\varphi$ is defined by the grammar:
\[
\varphi ::= a \mid \bot \mid \varphi_1 \otimes \varphi_2 \mid \previous\varphi \mid \varphi_1 \since \varphi_2 \mid \varphi_1 \trigger \varphi_2 \mid \Next \varphi \mid \varphi_1 \until \varphi_2 \mid \varphi_1 \release \varphi_2
\]
where $a\in\mathcal{A}$ is an atom and $\otimes$ is any binary Boolean connective $\otimes \in \{\to,\wedge,\vee\}$.
The last six cases correspond to the temporal connectives whose names are listed below:
\[
\begin{array}{r|cl}
\mathit{Past} & \previous & \text{for \emph{previous}}\\
              & \since    & \text{for \emph{since}}   \\
              & \trigger  & \text{for \emph{trigger}} \\
\end{array}
\qquad\qquad\qquad
\begin{array}{r|cl}
\mathit{Future} & \Next    & \text{for \emph{next}}\\
                & \until   & \text{for \emph{until}}\\
                & \release & \text{for \emph{release}}\\
\end{array}
\]
We also define several derived operators like the Boolean connectives
\(
\top \eqdef \neg \bot
\),
\(
\neg \varphi \eqdef  \varphi \to \bot
\),
\(
\varphi \leftrightarrow \psi \eqdef (\varphi \to \psi) \wedge (\psi \to \varphi)
\),
and the following temporal operators:
\[
\begin{array}{rcll}
       \alwaysP \varphi  & \eqdef & \bot \trigger \varphi             & \text{\emph{always before}} \\
   \eventuallyP \varphi  & \eqdef & \top \since \varphi               & \text{\emph{eventually before}} \\
             \initially\,& \eqdef & \neg \previous \top               & \text{\emph{initial}}\\
     \wprevious \varphi  & \eqdef & \previous \varphi \vee \initially & \text{\emph{weak previous}}
\end{array}
\qquad
\begin{array}{rcll}
       \alwaysF \varphi  & \eqdef & \bot \release \varphi             & \text{\emph{always afterward}}\\
   \eventuallyF \varphi  & \eqdef & \top \until \varphi               & \text{\emph{eventually afterward}}\\
               \finally  & \eqdef & \neg \Next \top                   & \text{\emph{final}}\\
         \wnext \varphi  & \eqdef & \Next \varphi \vee \finally       & \text{\emph{weak next}}
\end{array}
\]
As an example of a temporal formula, take for instance:
\begin{align}\label{f:fail}
\alwaysF ( \shoot \wedge \previous \eventuallyP \shoot \wedge \alwaysP \unloaded &\to \eventuallyF \fail)
\end{align}
capturing the sentence: {\em ``If we make two shots with a gun that was never loaded, then it will eventually fail.''}
Intuitively, `\alwaysF' is used to assert that the rule is applicable at every time point.
An explanation why shooting a loaded gun fails in unloading it,
could then be queried as follows
(where double negation allows us to express an integrity constraint).
\begin{align}\label{f:goal}
\alwaysF(\finally\to\neg\neg( \shoot \wedge \previous \loaded \wedge \loaded))
\end{align}

For the semantics, we start by defining a \emph{trace} of length $\lambda$ over alphabet $\mathcal{A}$ as a sequence $\tuple{H_i}_{i=0}^{\lambda}$ of sets $H_i\subseteq\mathcal{A}$.
We say that the trace is \emph{infinite} if $\lambda=\omega$ and \emph{finite} otherwise, that is, $\lambda=n$ for some natural number $0\leq n < \omega$.
We let $i=j..k$ stand for $i \in \mathbb{N}\cup\{\omega\}$
and $j \leq i \leq k$.
Given traces $\H=\tuple{H_i}_{i=0}^{\lambda}$ and $\H'=\tuple{H'_i}_{i=0}^{\lambda}$ both of length $\lambda$, we write $\H\leq\mathbf\H'$ if $H_i\subseteq H'_i$ for each $i=0..\lambda$;
accordingly, $\mathbf{H}<\mathbf{H'}$ iff both $\mathbf{H}\leq\mathbf{H'}$ and $\mathbf{H}\neq\mathbf{H'}$.

A \emph{Here-and-There trace} (for short \emph{\HT-trace}) of length $\lambda$ over alphabet $\mathcal{A}$ is a sequence of pairs $\tuple{H_i,T_i}_{i=0}^{\lambda}$ such that $H_i\subseteq T_i\subseteq \mathcal{A}$ for any $i=0..\lambda$.
As before, an \HT-trace is infinite if $\lambda=\omega$ and finite otherwise.
We often represent an \HT-trace as a pair of traces $\tuple{\H,\T}$ of length $\lambda$
where $\H=\tuple{H_i}_{i=0}^\lambda$ and $\T=\tuple{T_i}_{i=0}^\lambda$ and $\H \leq \T$.
\begin{definition}[Satisfaction]\label{def:tht:satisfaction}
  An \HT-trace $\tuple{\H,\T}$ of length $\lambda$ over alphabet $\mathcal{A}$ \emph{satisfies} a temporal formula $\varphi$ at time point $k=0..\lambda$, $k \neq \omega$,
  written $\M,k \models \varphi$, if the following conditions hold:
 \begin{enumerate}
  \item $\M, k \not\models \bot$
  \item $\M, k \models a$
    iff $a \in H_k$, for any atom $a \in \mathcal{A}$
  \item $\M, k \models \varphi \wedge \psi$
    iff
    $\M, k \models \varphi$
    and
    $\M, k \models \psi$
  \item
    $\M, k \models \varphi \vee \psi$
    iff
    $\M, k \models \varphi$
    or
    $\M, k \models \psi$
  \item
    $\M, k \models \varphi \to \psi$
    iff
    $\langle \mathbf{H}', \mathbf{T} \rangle, k \not \models \varphi$
    or
    $\langle \mathbf{H}', \mathbf{T} \rangle, k \models  \psi$, for all $\mathbf{H'} \in \{ \mathbf{H}, \mathbf{T} \}$
  \item $\M, k \models \previous \varphi$
    iff
    $k>0$ and $\M, k{-}1 \models \varphi$
  \item $\M, k \models \varphi \since \psi$
    iff
    for some $j=0..k$, we have
    $\M, j \models \psi$
    and
    $\M, i \models \varphi$ for all $i=j{+}1..k$
  \item $\M, k \models \varphi \trigger \psi$
    iff
    for all $j=0..k$, we have
    $\M, j \models \psi$
    or
    $\M, i \models \varphi$ for some $i=j{+}1..k$
  \item $\M, k \models \Next \varphi $
    iff
    $k<\lambda$ and $\M, k{+}1 \models \varphi$
  \item $\M, k \models \varphi \until \psi$
    iff
    for some $j=k..\lambda$, we have
    $\M, j \models \psi$
    and
    $\M, i \models \varphi$ for all $i=k..j{-}1$
  \item $\M, k \models \varphi \release \psi$
    iff
    for all $j=k..\lambda$, we have
    $\M, j \models \psi$
    or
    $\M, i \models \varphi$ for some $i=k..j{-}1$.\qed
  \end{enumerate}
\end{definition}
A formula $\varphi$ is a \emph{tautology}, written $\models \varphi$, iff $\M,k \models \varphi$ for any \HT-trace and any $k=0..\lambda$.
We call the logic induced by the set of all tautologies \emph{Temporal logic of Here and There} (\THT{} for short).
We say that an \HT-trace $\M$ is a \emph{model} of a set of formulas (or \emph{theory}) $\Gamma$ iff $\M,0 \models \varphi$ for any $\varphi \in \Gamma$.

When compared to standard temporal logics, the main peculiarity of Definition~\ref{def:tht:satisfaction} is the satisfaction of
implication $\varphi \to \psi$ that requires that both
(i) $\tuple{\H,\T},k\models \varphi$ implies $\tuple{\H,\T},k \models \psi$ and
(ii) $\tuple{\T,\T},k\models \varphi$ implies $\tuple{\T,\T},k \models \psi$.
This interpretation of `$\to$' is inherited from the (non-temporal) logic of Here and There~(\HT;~\citeNP{heyting30a}),
an intermediate logic dealing with exactly two worlds $\{h,t\}$ with the accessibility relation $h \leq t$
plus the reflexive closure $h \leq h$ and $t \leq t$.
This logic is weaker than classical logic and does not satisfy, among others, some classical tautologies such as the law of the \emph{Excluded Middle} $\varphi \vee \neg \varphi$.
A particular type of \HT-traces are the ones of form $\tuple{\T,\T}$ (that is, $\H=\T$) which we call \emph{total}.
Total models can be forced by adding the following variant of the excluded middle axiom schema:
\begin{align}
\alwaysF (a \vee \neg a) \tag{\EM}
\qquad
\text{ for each atom }
a \in \mathcal{A}
\text{ in the alphabet.}
\end{align}
Under total models, implication collapses to material implication and
\THT{} satisfaction $\tuple{\T,\T},k \models \varphi$ collapses to $\T,k\models \varphi$ in \LTL{} (for possibly infinite traces).
This implies that all \THT{} tautologies are \LTL{} tautologies but not vice versa, e.g.\ (\EM).

Another important remark is that the finiteness of $\M$ only affects the last three items of Definition~\ref{def:tht:satisfaction}
dealing with future-time operators.
In particular, if $\M$ has some finite length $\lambda=n$, then in the semantics for $\until$ and $\release$ (the last two items) $j$ ranges in the finite interval $\{k,\dots,n\}$.
Besides, if $\lambda=n$ the satisfaction of $\Next \varphi$ forces $k<n$ so that it implies that \emph{there does exist a next state} $k{+}1$.
As a result, the formula $\Next \top$ is not always satisfied, since it is false when $k=n=\lambda$.
On the other hand, when $\lambda=\omega$, time points $j$ in the semantics for $\until$ and $\release$ are just required to be $j\geq k$
without an upper limit.
Similarly, the condition $k<\lambda=\omega$ in the satisfaction of $\Next \varphi$ becomes obviously true,
and so irrelevant (a state $k{+}1$ is always granted).
%

The semantics for derived operators can also be easily deduced:
\begin{enumerate}
\setcounter{enumi}{11}
\item $\M, k \models \top$
\item $\M, k \models \alwaysP\varphi$
  iff
  $\M, i \models \varphi$ for all $i=0..k$
\item $\M, k \models \eventuallyP\varphi $
  iff
  $\M, i \models \varphi$ for some $i=0..k$
\item $\M, k \models \initially$
  iff
  $k =0$
\item $\M, k \models \wprevious\varphi$
  iff
  $k =0$ or
  $\M, k{-}1 \models \varphi$
\item $\M, k \models \alwaysF\varphi$
  iff
  $\M, i \models \varphi$ for any $i=k..\lambda$
\item $\M, k \models \eventuallyF\varphi$
  iff
  $\M, i \models \varphi$ for some $i=k..\lambda$
\item $\M, k \models \finally$
  iff $k=\lambda$
\item $\M, k \models \wnext\varphi$
  iff
  $k=\lambda$ or
  $\M, k{+}1 \models \varphi$
\end{enumerate}
We see that operators $\initially$ and $\finally$ exclusively depend on the value of time point $k$, so that the valuation for atoms from $\tuple{\H,\T}$ becomes irrelevant for them.
As a result, they behave ``classically'' and satisfy the excluded middle, that is,
$\models \initially \vee \neg \initially$ and $\models \finally \vee \neg \finally$ are \THT{} tautologies.
Besides, operator \finally{} has the additional peculiarity that it can only be true with finite traces (remember that $k \neq \omega$ in Def.~\ref{def:tht:satisfaction}).
This implies that the inclusion of axiom $\eventuallyF \finally$ in any theory forces its models to be finite traces, while including its negation
$\neg \eventuallyF \finally$ instead causes the opposite effect, that is, all models of the theory are infinite traces.

In this paper,
we consider several logics that are stronger than \THT{} and that can be obtained by the addition of axioms (or the corresponding restriction on sets of traces).
For instance, we define \THTo{} as $\THT+\{\neg \eventuallyF \finally\}$, that is, \THT{} where we exclusively consider infinite \HT-traces.%
\footnote{This corresponds to the (stronger) version of \THT{} considered previously by~\citeN{agcadipevi13a}.}
The finite-trace version, we call \THTf, corresponds to $\THT+\{\eventuallyF \finally\}$ instead.
Linear Temporal Logic for possibly infinite traces, \LTL, can be obtained as $\THT+\{(\EM)\}$, that is, \THT{} with total \HT-traces.
Accordingly, we can define \LTLo{} as $\THTo+\{(\EM)\}$, i.e.\ infinite and total \HT-traces, and obtain \LTLf{} as $\THTf+\{(\EM)\}$,
that is, \LTL{} on finite traces~\cite{giavar13a}.

In the rest of the paper,
we study several transformations preserving some kind of \THT-equivalence.
In this sense, it is important to observe that being equivalent is something generally stronger than simply having the same set of models.
We say that two formulas $\varphi, \psi$ are (globally) \emph{equivalent}, written $\varphi \gequiv \psi$, iff $\models \varphi \leftrightarrow \psi$, that is, $\M,k \models \varphi \leftrightarrow \psi$ for any \HT-trace $\M$ of length $\lambda$ and any $k=0..\lambda$, $k\neq \omega$.
Whenever $\varphi$ and $\psi$ are equivalent,
they are completely interchangeable when occurring in any theory $\Gamma$, without altering the semantic interpretation of $\Gamma$.
We say that $\varphi, \psi$ are just \emph{initially equivalent}, written $\varphi \iequiv \psi$, if they have the same models,
that is, $\M,0 \models \varphi$ iff $\M,0 \models \psi$, for any \HT-trace $\M$.
Obviously, $\varphi \gequiv \psi$ implies $\varphi \iequiv \psi$ but not vice versa.
To put a simple example, note that $\previous a \iequiv \bot$, since $\previous a$ is always false at the initial situation,
whereas in the general case $\previous a \not\gequiv \bot$ or, otherwise, we could always replace $\previous a$ by $\bot$ in any context.
The following are some generally useful properties satisfied in \THT.
\begin{proposition}[Persistence]\label{prop:persistance}
  Let $\tuple{\H,\T}$ be an \HT-trace of length $\lambda$ and $\varphi$ be a temporal formula.
  Then, for any $k=0..\lambda$, $k \neq \omega$,
  if $\tuple{\H,\T}, k \models \varphi$ then $\tuple{\T,\T}, k \models \varphi$ (or, if preferred, $\T,k \models \varphi$).\qed
\end{proposition}
As a corollary, we have that
$\langle\mathbf{H},\mathbf{T}\rangle \models \neg \varphi$ iff ${\T} \not\models \varphi$ in \LTL{}.
As said before, all \THT{} tautologies are \LTL{} tautologies but not vice versa.
However, they coincide for some types of equivalences, as stated below.
\begin{proposition}\label{prop:nonimpl}
  Let $\varphi$ and $\psi$ be formulas without implications (and so, without negations either). Then, $\varphi \equiv \psi$ in \LTL{} iff $\varphi \equiv \psi$ in \THT{}.
\end{proposition}
As an example, the usual inductive definition of the until operator from \LTL{}:
\begin{align}
  \varphi \until \psi & \equiv \psi \vee ( \varphi \wedge \Next (\varphi \until \psi) )\label{f:induntil}
\end{align}
is also valid in \THT{} due to Proposition~\ref{prop:nonimpl}.
In fact, we can exploit this result further.
By De Morgan laws, \LTL{} satisfies a kind of duality guaranteeing, for instance, that \eqref{f:induntil} iff:
\begin{align}
  \varphi \release \psi & \equiv  \psi \wedge ( \varphi \vee \wnext (\varphi \release \psi) ) \label{f:indrelease}
\end{align}
and, by Proposition~\ref{prop:nonimpl} again, this is also a valid equivalence in \THT{}.
Let us define all the pairs of dual connectives as follows: $\wedge/\vee$, $\top/\bot$, $\until/\release$, $\Next/\wnext$, $\alwaysF/\eventuallyF$, $\since/\trigger$, $\previous/\wprevious$, $\alwaysP/\eventuallyP$.
For a formula $\varphi$ without implications, we define $\delta(\varphi)$ as the result of replacing each connective by its dual operator.
Then, we get the following corollary of Proposition~\ref{prop:nonimpl}.
%
\begin{corollary}[Boolean Duality]\label{BDT}
  Let $\varphi$ and $\psi$ be formulas without implication. Then, \THT{} satisfies: $\varphi \gequiv \psi$  iff $\delta(\varphi) \gequiv \delta(\psi)$.
\end{corollary}

In a similar manner, we can also exploit the temporal symmetry in the system so we can switch the temporal direction of operators to conclude, for instance, that \eqref{f:induntil} iff $\varphi \since \psi \equiv \psi \vee ( \varphi \wedge \previous (\varphi \since \psi) )$.
This second type of duality, however, has some obvious limitations when we allow for infinite traces.
In that case, for instance, the past has a beginning $\eventuallyP \initially \equiv \top$ but the future may have no end $\eventuallyF \finally \not\equiv \top$.
If we restrict ourselves to finite traces, we get the following result.
Let $\until/\since$, $\release/\trigger$, $\Next/\previous$, $\wnext/\wprevious$, $\alwaysF/\alwaysP$, and $\eventuallyF/\eventuallyP$ denote all pairs of swapped-time connectives and let $\sigma(\varphi)$ denote the replacement in $\varphi$ of each connective by its swapped-time version.
\begin{lemma}\label{TDT}
There exists a mapping $\varrho$ on finite \THT-traces of the same length $n\geq 0$ such that for any $k=0..n$,
$\M,k \models \varphi$ if and only if $\varrho(\M),n\!-\!k \models \sigma(\varphi)$.
\end{lemma}%
\begin{theorem}[Temporal Duality Theorem]
  A temporal formula $\varphi$ is a $\THTf$-tautology if and only if $\sigma(\varphi)$ is a $\THTf$-tautology.
\end{theorem}
%

Before introducing non-monotonicity, we begin by providing notation for representing sets of \THT{} models.
We write $\THT(\Gamma,\lambda)$ to stand for the set of \THT{} models of length $\lambda$ of a theory $\Gamma$,
and define $\THT(\Gamma) \eqdef \bigcup_{\lambda=0}^\omega \THT(\Gamma,\lambda)$, that is, the whole set of models of $\Gamma$ of any length.
Given a set of \THT{} models, we define the ones in equilibrium as follows.
\begin{definition}[Temporal Equilibrium Model]
Let $\mathfrak{S}$ be some set of \HT-traces.
A total \HT-trace $\tuple{\T,\T} \in\mathfrak{S}$ is a \emph{temporal equilibrium model} of $\mathfrak{S}$ iff
there is no other $\H < \T$ such that $\tuple{\H,\T} \in\mathfrak{S}$.
\qed
\end{definition}
%
If this is the case, we also say that trace \T{} is a \emph{temporal stable model} of $\mathfrak{S}$.
We further talk about temporal equilibrium or temporal stable models of a theory $\Gamma$ when $\mathfrak{S}=\THT(\Gamma)$, respectively.
Moreover, we write $\TEL(\Gamma,\lambda)$ and $\TEL(\Gamma)$ to stand for the temporal equilibrium models of $\THT(\Gamma,\lambda)$ and $\THT(\Gamma)$ respectively.

Since the ordering relation among traces is only defined for a fixed $\lambda$, the following can be easily observed:
\begin{proposition}
The set of temporal equilibrium models of $\Gamma$ can be partitioned by the trace length $\lambda$, that is,
$\bigcup_{\lambda=0}^\omega \TEL(\Gamma,\lambda) = \TEL(\Gamma)$. \qed
\end{proposition}

\emph{Temporal Equilibrium Logic} (\TEL) is the (non-monotonic) logic induced by temporal equilibrium models.
We can also define the variants \TELo{} and \TELf{} by applying the corresponding restriction to infinite or finite traces, respectively.

As an example of non-monotonicity,
consider the formula
\begin{align}\label{f:ys4b}
 \alwaysF (\previous \loaded \wedge \neg \unloaded & \to  \loaded)
\end{align}
along with literal \loaded{} which combines the inertia for \loaded{} with the initial state for that fluent.
Without entering into much detail, this formula behaves as the logic program consisting of fact \lstinline{loaded(0)} and rule
`\lstinline{loaded(T) :- loaded(T-1), not unloaded(T)}'
for any time point \lstinline{T>0}.
As expected, for some fixed $\lambda$, we get a unique temporal stable model of the form \T{} such that $T_i=\{\loaded\}$ for $i=0..\lambda$,
as there is no reason to become \unloaded.
Note that in the most general case of \TEL, we actually get one stable model per each possible $\lambda$, including $\lambda=\omega$.
Now, consider formula \eqref{f:ys4b} with $\loaded \wedge \Next \Next \unloaded$ which amounts to adding the fact \lstinline{unloaded(2)}.
As expected, for each $\lambda$,
the only temporal stable model now is $T_0=T_1=\{\loaded\}$, $T_2=\{\unloaded\}$ and $T_i=\emptyset$ for $i=3..\lambda$.
Note that by making $\Next \Next \unloaded$ true we are also forcing $\lambda \geq 3$, that is, there are no temporal stable models (nor even \THT{} models) of length smaller than three.
Thus, by adding the new information $\Next \Next \unloaded$ some conclusions that could be derived before, such as $\alwaysF \loaded$, are not derived any more.

As an example emphasizing the behavior of finite traces, take the formula $\alwaysF (\neg a \to \Next a)$ which can be seen as a program rule
`\lstinline{a(T+1) :- not a(T)}' for any natural number \lstinline{T}.
As expected, temporal stable models make $a$ false in even states and true in odd ones.
However, we cannot take finite traces where the final state $\lambda$ makes $a$ false, since the rule would force $\Next a$ and this implies the existence of a successor state.
As a result, the temporal stable models of this formula have the (regular expression) form $(\emptyset \ \{a\})^+$ for finite traces in $\TELf$, or the infinite trace $(\emptyset \ \{a\})^\omega$ in \TELo.

Another interesting example is the temporal formula $\alwaysF (\neg \Next a \to a) \wedge \alwaysF (\Next a \to a)$.
The corresponding rules
`\lstinline{a(T) :- not a(T+1)}'
and
`\lstinline{a(T) :- a(T+1)}'
have no stable model~\cite{fages94a} when grounded for all natural numbers \lstinline{T}.
This is because there is no way to build a finite proof for any \lstinline{a(T)}, as it depends on infinitely many next states to be evaluated.
The same happens in \TELo, that is, we get no temporal stable model, but in \TELf, we can use the fact that $\Next a$ is always false in the last state.
Then, $\alwaysF(\neg \Next a \to a)$ supports $a$ in that state and therewith $\alwaysF (\Next a \to a)$ inductively supports $a$ everywhere.

As an example of a temporal expression not so close to logic programming consider, for instance, the formula $\alwaysF\eventuallyF a$, which is normally used in \LTLo{} to assert that $a$ occurs infinitely often.
As discussed in~\cite{giavar13a}, if we assume finite traces, then the formula collapses to $\alwaysF (\finally \to a)$ in \LTLf,
that is, $a$ is true at the final state (and true or false everywhere else).
The same behavior is obtained in \THTo{} and \THTf, respectively.
However, if we move to \TEL, a truth minimization is additionally required.
As a result, in \TELf{} for a fixed $\lambda \in \mathbb{N}$,
we obtain a unique temporal stable model where $a$ is true at the last state, and false everywhere else,
whereas \TELo{} yields no temporal stable model at all.
This is because for any $\T$ with an infinite number of $a$'s we can always take some $\H$ where we remove $a$ at one state, and still have an infinite number of $a$'s in $\H$.
Thus, for any total \THTo{} model $\tuple{\T,\T}$ of $\alwaysF \eventuallyF a$ there always exists some model $\tuple{\H,\T}$ with strictly smaller $\H < \T$.



\section{Temporal Logic Programs}
\label{sec:lpnf}

%
Our computational approach to \TELf{} relies on a reduction to a normal form suitable for ASP systems.
For this,
we identified the class of temporal logic programs defined next.
%
\begin{definition}[Temporal literal, rule, and program]\label{def:temporal:rule}
Given alphabet $\mathcal{A}$, we define the set of \emph{temporal literals} as
$\{a, \neg a, \previous a, \neg \previous  a \mid a \in \mathcal{A}\}$.

A \emph{temporal rule} is either:\\
\begin{tabular}{@{\hspace{-120pt}}lr@{\hspace{-4pt}}l}
-- \ an \emph{initial rule} of the form &                           $\Bd$ & $\to \Hd$     \\
-- \ a  \emph{dynamic rule} of the form &  $\wnext\,\alwaysF (       \Bd$ & $\to \Hd)$    \\
-- \ a  \emph{final   rule} of the form &  $\alwaysF(\,\finally \to (\Bd$ & $\to \Hd )\,)$
\end{tabular}\\
where $\Bd=b_1 \wedge \dots \wedge b_n$ with $n\geq 0$, $\Hd=a_1 \vee \dots \vee a_m$
with $m \geq 0$ and the $b_i$ and $a_j$ are temporal literals for dynamic rules,
or regular literals $\{a, \neg a \mid a \in \mathcal{A}\}$ for initial and final rules.
A \emph{temporal logic program} (\TELf{} \emph{program}, for short) is a set of temporal rules.
\end{definition}
%
We let \initial{P}, \dynamic{P}, and \final{P} stand for the set of all initial, dynamic, and final rules in a \TELf{} program $P$, respectively.
A \TELf{} program just consisting of initial rules amounts to a regular logic program.
%
%
Dynamic rules are preceded by the weak version of next \wnext{} rather than \Next{}
since we deal with finite traces and the final state has no subsequent state.

Our earlier examples in \eqref{f:goal} and \eqref{f:ys4b} are already close to \TELf{} programs,
and a minor transformation yields the following temporal rules equivalent to \eqref{f:goal} and \eqref{f:ys4b}, respectively.
\begin{align*}
  \wnext\,\alwaysF( \shoot \wedge \previous \loaded \wedge \loaded\to\goal)
  &\wedge
  \alwaysF (\finally \to ( \neg \goal \to \bot))
  \\
  \wnext\,\alwaysF (\previous \loaded \wedge \neg \unloaded \to  \loaded)
  &
\end{align*}
Note that no initial rules are needed since $\previous \loaded$ is false at the initial time point,
and \goal{} is a new auxiliary atom.
In the remainder, for illustration purposes, we use the simple \TELf{} program $P$:
\begin{align}
\{\
                                                 \to a      ,\quad
\label{ex:program} \wnext\,\alwaysF (\previous a \to b)     ,\quad
                   \alwaysF (\finally\to (\neg b \to\bot))
\ \}
\end{align}
which has a single finite temporal stable model of length 1, viz.\ $\langle\{a\},\{b\}\rangle$.

The following result warrants that \TELf{} programs constitute indeed a normal form.
%
\begin{theorem}[Normal form]\label{thm:normalform}
Every temporal formula $\varphi$ can be converted into a \TELf{} program \THTf-equivalent to $\varphi$.
\end{theorem}
%
For transforming arbitrary temporal formulas into normal form,
we use a Tseitin-style reduction~\citeyear{tseitin68a} that relies on an alphabet extended
by new atoms for each formula in the original language.
The equivalence result in Theorem~\ref{thm:normalform} is then obtained after removing auxiliary atoms and,
in fact, is still preserved inside the context of a larger theory for the original vocabulary
(i.e.\ we have \emph{strong equivalence} modulo auxiliary atoms).

%
Now, given a \TELf{} program $P$ and a fixed (finite) $\lambda$, we can compute all models in $\TEL(P,\lambda)$ by a translation of $P$ into a regular program.
For this, we let
\(
\mathcal{A}_k=\{\Stamp{a}{k}\mid a\in\mathcal{A}\}
\)
be a time stamped copy of alphabet $\mathcal{A}$ for each time point $k=0..\lambda$.
%
\begin{definition}[Bounded translation]\label{def:temporal:bounded:translation}
  We define the translation $\tau$ of a temporal literal at time point $k$ as
  \begin{align*}
    \tau_k(          a) &\eqdef  \Stamp{a}{k    } &\tau_k(\neg            a) &\eqdef \Stamp{\neg a}{k    } & &\text{ for }a\in\mathcal{A}\\
    \tau_k(\previous a) &\eqdef \Stamp{a}{k{-}1} &\tau_k(\neg \previous  a) &\eqdef \Stamp{\neg a}{k{-}1} & &\text{ for }a\in\mathcal{A}
  \end{align*}
  We define the translation of any temporal rule $r$ in Definition~\ref{def:temporal:rule} at time point $k$ as
  \begin{align*}
    \tau_k(r) &\eqdef \tau_k(a_1) \vee \dots \vee \tau_k(a_m)\leftarrow\tau_k(b_1) \wedge \dots \wedge \tau_k(b_n)
  \end{align*}
  We define the translation of a temporal program $P$ bounded by finite length $\lambda$ as
  \[\textstyle
  \tau_\lambda(P)\eqdef \{ \tau_0(r)\mid r\in\initial{P}\}\ \cup \ \{\tau_k(r)\mid r\in\dynamic{P}, k=1..\lambda\}\ \cup\ \{\tau_\lambda(r)\mid r\in\final{P}\}
  \]
\end{definition}
%
%
Note that the translation of temporal rules is similar in just considering the implication $B \to A$ in Definition~\ref{def:temporal:rule};
their difference manifests itself in their instantiation in $\tau_\lambda(P)$.

Applying translation $\tau$ for some bound $\lambda$ to our \TELf{} program $P$ in \eqref{ex:program}
yields regular logic programs of the following form.
\[
\tau_\lambda(P)=
\{ a_0 \leftarrow {} \}
\ \cup\
\{ b_k \leftarrow a_{k-1}\mid k=1..\lambda\}
\ \cup\
\{ \bot  \leftarrow \neg b_\lambda \}
\]
Program $\tau_1(P)$ has the stable model $\{a_0,b_1\}$, but all $\tau_\lambda(P)$ for $\lambda>1$ are unsatisfiable.

\begin{theorem}\label{thm:temporal:bounded:translation}
Let $P$ be a \TELf{} program over $\mathcal{A}$.
Let $\T=\tuple{T_i}_{i=0}^{\lambda}$ be a trace of finite length $\lambda$ over $\mathcal{A}$
and
$X$ a set of atoms over $\bigcup_{0\leq i\leq \lambda}\mathcal{A}_i$
such that
\(
a\in T_i\text{ iff }a_i\in X\text{ for }0\leq i\leq \lambda
\).

Then,
\T{} is a temporal stable model of $P$
iff
$X$ is a stable model of $\tau_\lambda(P)$.
\end{theorem}
%
Applied to our example,
this result confirms that the temporal stable model $\langle\{a\},\{b\}\rangle$ of $P$
corresponds to the stable model $\{a_0,b_1\}$ of $\tau_2(P)$.

Using this translation we have implemented a system, \tel\footnote{\url{https://github.com/potassco/tel}}, that
takes a propositional theory $\Gamma$ of arbitrary \TELf{} formulas and a bound $\lambda$ and returns the regular logic program $\tau_\lambda(P)$,
where $P$ is the intermediate normal form of $\Gamma$ left implicit.
The resulting program $\tau_\lambda(P)$ can then be solved by any off-the-shelf ASP system.
For illustration,
consider the representation of our example temporal program in~\eqref{ex:program} in \tel's input language.
\begin{lstlisting}[numbers=none,belowskip=2pt,aboveskip=2pt,basicstyle=\ttfamily]
  a.
  #next^ #always+ ( (#previous a) -> b).
  #always+ ( #final -> (~ b -> #false)).
\end{lstlisting}
As expected, passing the result of \tel's translation for horizon~1 to \clingo{} yields the stable model containing \lstinline{a(0)} and \lstinline{b(1)}
(suppressing auxiliary atoms).

%
The bounded translation $\tau_\lambda(P)$ allows us to compute all models in $\TEL(P,\lambda)$ for a fixed bound $\lambda$.
However, in many practical problems (as in planning, for instance), $\lambda$ is unknown beforehand and
the crucial task consists in finding a representation of $\TEL(P,k)$ that is easily obtained from that of $\TEL(P,k\text{-}1)$.
In ASP, this can be accomplished via incremental solving techniques that rely upon the composition of logic program modules~\cite{oikjan06a}.
The idea is then to associate the knowledge at each time point with a module and to successively add modules corresponding to increasing time points
(while leaving all previous modules unchanged).
A stable model obtained after $k$ compositions then corresponds to a \TELf{} model of length $k$.
This technique of modular computation, however, is only applicable when modules are \emph{compositional} (positive loops cannot be formed across modules), something that cannot always be guaranteed for arbitrary \TELf{} programs.
Still, we identify a quite general syntactic fragment\footnote{In order to compute loop formulas for \TELo{}, \cite{cabdie11a} used a similar fragment (\emph{splittable programs}) where rules cannot derive information from the future to the past.} that implies compositionality.
We say that a temporal rule as in Definition~\ref{def:temporal:rule} is \emph{present-centered},
whenever all the literals $a_1,\dots,a_m$ in its head $\Hd$ are regular.
Accordingly, a set of such rules is a present-centered \TELf{} program.
In fact,
such programs are sufficient to capture common action languages,
as witnessed by the correspondence between dynamic temporal rules and static and dynamic laws in action language ${\cal BC}$~\cite{leliya13a}:%
\footnote{In $\mathcal{BC}$, \textbf{ifcons} stands for ``if consistent'', while \textbf{if} and \textbf{after} have their literal meaning.}
\[
  \begin{array}{ll}
    a\textbf{ if }   b_1,\dots,b_m\textbf{ ifcons } c_1,\dots,c_n & \wnext\,\alwaysF(           b_1\wedge\dots\wedge          b_m\to a\vee\neg c_1\vee\dots\vee \neg c_n)\\
    & \ \ \wedge \ (b_1\wedge\dots\wedge          b_m\to a\vee\neg c_1\vee\dots\vee \neg c_n)
    \\
    a\textbf{ after }b_1,\dots,b_m\textbf{ ifcons } c_1,\dots,c_n & \wnext\,\alwaysF( \previous b_1\wedge\dots\wedge\previous b_m\to a\vee\neg c_1\vee\dots\vee \neg c_n)
  \end{array}
\]

Following these ideas, we provide next a ``\emph{point-wise}'' variant of our translation that allows for defining one module per time point and is compositional for the case of present-centered \TELf{} programs.
We begin with some definitions.
A \emph{module}~\module{P} is a triple
\(
(P,I,O)
\)
consisting of
a logic program~$P$ over alphabet $\mathcal{A}_P$
and sets~$I$ and~$O$
of \emph{input} and \emph{output}
atoms such that
(i) $I\cap\nolinebreak O=\nolinebreak\emptyset$,
(ii) $\mathcal{A}_P\subseteq I\cup O$, and
(iii) $\Head{P}\subseteq O$,
where \Head{P} gives all atoms occurring in rule heads in $P$.
Whenever clear from context, we associate \module{P} with $(P,I,O)$.
In our setting,
a set $X$ of atoms is a stable model of \module{P},
if $X$ is a stable model of logic program
\(
P 
\).%
\footnote{Note that the default value assigned to input atoms is \emph{false} in multi-shot solving~\cite{gekakasc17a};
  this differs from the original definition~\cite{oikjan06a} where a choice rule is used.}
Two modules~$\module{P}_1$ and~$\module{P}_2$ are
\emph{compositional}, if
$O_1\cap O_2=\emptyset$
and
$O_1\cap C=\emptyset$ or $O_2\cap C=\emptyset$
for every strongly connected component~$C$ of the positive dependency graph of the logic program $P_1\cup P_2$.
In other words,
all rules defining an atom must belong to the same module, and no positive recursion is allowed among modules.
Whenever $\module{P}_1$ and~$\module{P}_2$ are compositional,
their \emph{join} is defined as the module
\(
\module{P}_1\sqcup\module{P}_2
=
(P_1\cup P_2,(I_1\setminus O_2)\cup (I_2\setminus O_1), O_1\cup O_2)
\).
The module theorem~\cite{oikjan06a} ensures that compatible stable models of $\module{P}_1$ and~$\module{P}_2$ can be combined to one of
$\module{P}_1\sqcup\module{P}_2$,
and vice versa.

For literals and rules, the point-wise translation $\tau^*$ coincides with $\tau$ up to final rules.
%
\begin{definition}[Point-wise translation: Temporal rules]\label{def:temporal:pointwise:translation}
  %
  We define the translation of a final rule $r$ as in Definition~\ref{def:temporal:rule} at time point $k$ as
  \begin{align}\label{eq:temporal:pointwise:translation:tri}
    \tau^*_k(r) &\eqdef \tau_k(a_1) \vee \dots \vee \tau_k(a_m)\leftarrow\tau_k(b_1) \wedge \dots \wedge \tau_k(b_n)\wedge\neg\Stamp{q}{k+1}
  \end{align}
  for a new atom $q\notin{\mathcal{A}}$
  and of an initial or dynamic rule $r$ as $\tau^*_k(r)\eqdef\tau_k(r)$.
\end{definition}
%
The new atoms $q_{k+1}$ in \eqref{eq:temporal:pointwise:translation:tri} are used to deactivate instances of final rules.
This allows us to implement operator \finally{} by using $\neg q_{k+1}$ and therefore to enable the actual final rule unless $q_{k+1}$ is derivable.
The idea is then to make sure that at each horizon $k$ the atom $q_{k+1}$ is false, while $q_1,\dots,q_k$ are true.
In this way, only $\tau^*_k(r)$ is potentially applicable, while all rules $\tau^*_i(r)$ are disabled at earlier time points $i=1..k{-}1$.

Translation $\tau^*$ is then used to define modules for each time point as follows.
%
\begin{definition}[Point-wise translation: Modules]\label{def:temporal:module}
Let $P$ be a present-centered \TELf{} program over $\mathcal{A}$.
We define the module $\module{P}_k$ corresponding to $P$ at time point $k$ as
\begin{align*}
  \module{P}_0 &\eqdef (P_0,                     \{\Stamp{q}{  1}\},{\mathcal{A}_0}                    )&
  \module{P}_k &\eqdef (P_k,\mathcal{A}_{k-1}\cup\{\Stamp{q}{k+1}\},{\mathcal{A}_k}\cup\{\Stamp{q}{k}\})
                  \quad\text{ for }k>0
\end{align*}
where
\begin{align*}
  P_0 &\eqdef \{\tau^*_0(r)\mid r\in\;\initial{P}\} \cup\{\tau^*_0(r)\mid r\in\final{P}\}
  \\
  P_k &\eqdef \{\tau^*_k(r)\mid r\in  \dynamic{P}\} \cup\{\tau^*_k(r)\mid r\in\final{P}\}\cup\{q_k \leftarrow \}
\end{align*}
\end{definition}
%
Each module $\module{P}_k$ defines what holds at time point $k$.
The underlying present-centeredness warrants that
modules only incorporate atoms from previous time points,
as reflected by $\mathcal{A}_{k-1}$ in the input of $\module{P}_k$.
The exception consists of
auxiliary atoms like $q_{k+1}$ that belong to the input of each $\module{P}_k$ for $k>0$
but only get defined in the next module $\module{P}_{k+1}$.
This corresponds to the aforementioned idea that $q_{k+1}$ is false when $\module{P}_k$ is the final module,
and is set permanently to true once the horizon is incremented by adding $\module{P}_{k+1}$.
Note that atoms like $q_{k+1}$ only occur negatively in rule bodies in $\module{P}_k$ and
hence cannot invalidate the modularity condition.
This technique allows us to capture the transience of final rules.

The point-wise translation of our present-centered example program $P$ from \eqref{ex:program} yields the following modules.
\begin{align*}
  \module{P}_0 & =
  \left(
    \{a_0\leftarrow{}\}\cup\{\leftarrow \neg b_0,\neg q_1\},
    \{q_1\},
    \{a_0,b_0\}
  \right)
  \\
  \module{P}_i & =
  \left(
    \{b_i\leftarrow a_{i-1}\}\cup\{\leftarrow \neg b_i,\neg q_{i+1}\}\cup \{q_i \leftarrow\},
    \{a_{i-1},b_{i-1},q_{i+1}\},
    \{a_i,b_i,q_i\}
  \right)
  \\\textstyle\hspace{-25pt}
  \bigsqcup_{i=0}^\lambda\module{P}_i
  &=\textstyle
  (
    P_0\cup\bigcup_{i=1}^\lambda P_i,
    \{q_{\lambda+1}\},
    \{a_i,b_i\mid i=0..\lambda\} \cup \{q_i\mid i=1..\lambda\}
    )
\end{align*}
As above, only the composed module for $\lambda=1$ yields a stable model, viz.\ $\{a_0,b_1,q_1\}$.

%
\begin{theorem}\label{thm:temporal:pointwise:translation}
Let $P$ be a present-centered \TELf{} program over $\mathcal{A}$.
Let $\T=\tuple{T_i}_{i=0}^{\lambda}$ be a trace of finite length $\lambda$ over $\mathcal{A}$
and
$X$ a set of atoms over $\bigcup_{0\leq i\leq \lambda}\mathcal{A}_i$
such that
\(
a\in T_i\text{ iff }a_i\in X\text{ for }0\leq i\leq \lambda
\).
Then,

\T{} is a temporal stable model of $P$
iff
$X\cup\{q_i \mid i=1..\lambda\}$ is a stable model of
\(
\bigsqcup_{i=0}^\lambda\module{P}_i.
\)
\end{theorem}
%
As with Theorem~\ref{thm:temporal:bounded:translation},
this result confirms that the temporal stable model $\langle\{a\},\{b\}\rangle$ of $P$
corresponds to the stable model $\{a_0,b_1,q_1\}$ of $\module{P}_0\sqcup\module{P}_1$.

As one might expect, not any \TELf{} theory is reducible to a present-centered \TELf{} program.
Hence, computing models via incremental solving imposes some limitations on the possible combinations of temporal operators.
Fortunately, we can identify again a quite natural and expressive syntactic fragment that is always reducible to present-centered programs.
We say that a temporal formula is a \emph{past-future rule} if it consists of rules as those in Definition~\ref{def:temporal:rule} where
$\Bd$ and $\Hd$ are just temporal formulas with the following restrictions:
$\Bd$ and $\Hd$ contain no implications other than negations ($\alpha \to \bot$),
$\Bd$ contains no future operators, and
$\Hd$ contains no past operators.
An example of a past-future rule is \eqref{f:fail}.
Then, we have the following result.
\begin{theorem}[Past-future reduction]\label{thm:normalform:pc}
Every past-future rule $r$ can be converted into a present-centered \TELf{} program that is \TELf{}-equivalent to $r$.
\end{theorem}

We have implemented a second system, \telingo\footnote{\url{https://github.com/potassco/telingo}}, that
deals with present-centered \TELf{} programs that are expressible in the full (non-ground) input language of \clingo{} extended with temporal operators.
In addition, \telingo{} offers several syntactic extensions to facilitate temporal modeling:
First, next operators can be used in singular heads and,
second, arbitrary temporal formulas can be used in integrity constraints.
All syntactic extensions beyond the normal form of Theorem~\ref{thm:normalform}
are compiled away by means of the translation used in its proof.
The resulting present-centered \TELf{} programs are then processed according the point-wise translation.

To facilitate the use of operators \previous{} and \Next{},
\telingo{} allows us to express them
by adding leading or trailing quotes to the predicate names of atoms, respectively.
For instance, the temporal literals $\previous p(a)$ and $\Next q(b)$ can be expressed by \lstinline{'p(a)} and \lstinline{q'(b)}, respectively.
%
%
For another example,
consider the representation of the sentence
\emph{``A robot cannot lift a box unless its capacity exceeds the box's weight plus that of all held objects''}:
\begin{lstlisting}[numbers=none,belowskip=2pt,aboveskip=2pt,basicstyle=\ttfamily]
 :- lift(R,B), robot(R), box(B,W),
    #sum { C : capacity(R,C); -V,O : 'holding(R,O,V) } < W.
\end{lstlisting}
Atom \lstinline{'holding(R,O,V)} expresses what the robot was holding at the \emph{previous} time point.

The distinction between different types of temporal rules is done in \telingo{} via \clingo's \lstinline{#program} directives~\cite{gekakasc17a},
which allow us to partition programs into subprograms.
More precisely,
each rule in \telingo's input language is associated with
a temporal rule $r$ of form $(b_1 \wedge \dots \wedge b_n \to a_1 \vee \dots \vee a_m)$ as in Definition~\ref{def:temporal:rule} and
interpreted as $r$, $\wnext\alwaysF{r}$, or $\alwaysF(\finally \to r)$ depending on whether it occurs
in the scope of a program declaration headed by \lstinline{initial}, \lstinline{dynamic}, or \lstinline{final}, respectively.
Additionally, \telingo{} offers \lstinline{always} for gathering rules preceded by \alwaysF{}
(thus dropping \wnext{} from dynamic rules).
A rule outside any such declaration is regarded to be in the scope of \lstinline{initial}.
This allows us to represent the \TELf{} program in~\eqref{ex:program} in the two alternative ways shown in Table~\ref{tab:enc:programs}.
%
\begin{table}[ht]
\centering
\begin{minipage}[t]{120pt}
\begin{lstlisting}[frame=single,numbers=none,belowskip=0pt,aboveskip=0pt,basicstyle=\ttfamily]
#program initial.
a.

#program dynamic.
b :- 'a.

#program final.
  :- not b.
\end{lstlisting}
\end{minipage}
\qquad\qquad
\begin{minipage}[t]{120pt}
\begin{lstlisting}[frame=single,numbers=none,belowskip=0pt,aboveskip=0pt,basicstyle=\ttfamily]
#program always.
a :- &initial.


b :- 'a.


  :- not b, &final.
\end{lstlisting}
\end{minipage}%
\caption{Two alternative \telingo{} encodings for the \TELf{} program in~\eqref{ex:program}}
\label{tab:enc:programs}
\end{table}

As mentioned,
\telingo{} allows us to use nested temporal formulas in integrity constraints as well as in negated form in place of temporal literals within rules.
This is accomplished by encapsulating temporal formulas like $\varphi$ in expressions of the form
`\lstinline[mathescape]+&tel { $\varphi$ }+'.
To this end, the full spectrum of temporal operators is at our disposal.
They are expressed by operators built from \lstinline{<} and \lstinline{>} depending on whether they refer to the past or the future, respectively.
So,
\lstinline{<}/1, \lstinline{<?}/2, and \lstinline{<*}/2
stand for \previous, \since, and \trigger,
and
\lstinline{>}/1, \lstinline{>?}/2, \lstinline{>*}/2
for \Next, \until, \release.
Accordingly,
\lstinline{<*}/1,
\lstinline{<?}/1,
\lstinline{<:}/1
represent \alwaysP, \eventuallyP, \wprevious, and
analogously their future counterparts.
\initially{} and \finally{} are  are represented by \lstinline{&initial} and \lstinline{&final}.
This is complemented by Boolean connectives \lstinline{&}, \lstinline{|}, \lstinline{~}, etc.
For example, the integrity constraint
`\(
\shoot \wedge \alwaysP \unloaded \wedge \previous \eventuallyP \shoot \to \bot
\)'
is expressed as follows.
\begin{lstlisting}[numbers=none,belowskip=2pt,aboveskip=2pt,basicstyle=\ttfamily]
   :- shoot, &tel { <* unloaded & < <? shoot }.
\end{lstlisting}

Once \telingo{} has translated an (extended) \TELf{} program into a regular one,
it is incrementally solved by \clingo's multi-shot solving engine~\cite{gekakasc17a}.


\section{Discussion and conclusions} 
\label{sec:conclusion}

For incorporating temporal representation and reasoning into existing ASP solving technology,
we introduced a variant of Temporal Equilibrium Logic, \TELf, that deals with finite traces,
something better aligned with incremental ASP solving for dynamic domains.
The original version of this logic, \TELo~\cite{agcadipevi13a}, was exclusively thought for infinite traces and,
accordingly, its computation~\cite{cabdie11a} is done in terms of automata obtained by a model checker.
This strategy is more adequate for checking properties of reactive systems
but not so convenient when looking for minimal plans, performing temporal explanation, or even for diagnosis on finite executions.
To analyze which logical properties may vary depending on the finiteness assumption,
we defined a more general (and weaker) version of \TEL{} and its monotonic basis \THT{}, which accepts both finite and infinite traces.
This general \TEL{} acts as an umbrella to study the relation of the new finite trace variants,
\TELf{} and \THTf{}, with their temporal predecessors \TELo{}, \LTLo, \LTLf{} as well as \HT{} and its equilibria.
For instance, we may conclude that satisfiability for both \THTf{} and \TELf{} are \textsc{PSpace}-hard,
since \LTLf{}, proved to be \textsc{PSpace}-complete \cite{giavar13a}, can be easily reduced to \THTf{} or \TELf{} by adding the excluded middle axiom (\EM).
In fact, \THTf{} is also \textsc{PSpace}-complete:
its membership can be proved by encoding \THTf{} into \LTLf{} as in the reduction from \THTo{} to \LTLo{} made in~\cite{cabdem11a}.
In the case of \TELf{} satisfiability, we conjecture its \textsc{ExpSpace} membership by using a translation into \TELo{},
which is \textsc{ExpSpace}-complete \cite{bozpea15a}, similar to the one from \LTLf{} into \LTLo{} in~\cite{giavar13a}.
A detailed complexity analysis remains future work.

As with \TELo~\cite{cabalar10a}, we proved that \TELf{} can be reduced to a normal form close to logic programs.
Moreover, the one for \TELf{} happens to be significantly simpler, since it does not need to resort to nested global operators. 
We developed two translations of this normal form into ASP:
(i) one to obtain temporal stable models of fixed length; and
(ii), another based on the composition of logic program modules,
allowing for incremental computation.
These translations gave rise to two different systems.
Our first system, \tel{}, accepts an arbitrary propositional \TELf-theory and a bound and then reduces it to normal form
followed by translation (i) into ASP.
This allows us to harness the full expressiveness of a temporal language
while using any off-the-shelf ASP system.
Our second system, \telingo{}, extends the ASP system \clingo{} to compute the temporal stable models of (non-ground) temporal logic programs.
To this end, it extends the full-fledged input language of \clingo{} with temporal operators
and computes temporal models incrementally by multi-shot solving~\cite{gekakasc17a} using translation (ii) into ASP.
It is also interesting to observe that \TELf{} sheds light on existing concepts used in incremental ASP solving,
when interpreting increments as time-points.
For instance, operator \finally{} naturally corresponds to the so-called ``external query atom''~\cite{gekakasc17a} used for progressing goal conditions,
while the syntactic form of present-centered programs reflects the modeling methodology~\cite{gekasc12a} put forward for incremental ASP solving that avoids ``future atoms'' referring to time point \texttt{T+1} in rule heads.

All in all, \TELf{} offers an expressive, semantically well founded language for modeling dynamic systems in ASP
that allows for exploiting existing solving technology and, at the same time,
enables a fully logical analysis of temporal properties,
either from plain ASP specifications or from action languages that can be naturally translated into \TELf.
For future work, we plan to investigate some topics already studied for \TELo{} in the case of finite traces,
such as characterizing strong equivalence, checking unsatisfiability by automata-based methods,
or improving the efficiency of grounding for temporal programs.


\paragraph{Acknowledgments.}
This work was partially supported by
MINECO, Spain, grant TIC2017-84453-P,
Xunta de Galicia, Spain (GPC ED431B 2016/035 and 2016-2019 ED431G/01, CITIC), and
DFG grant SCHA 550/9.


\bibliographystyle{acmtrans}


\end{document}